\pdfoutput=1

\documentclass[11pt]{article}

\usepackage[final]{acl}

\usepackage{times}
\usepackage{latexsym}
\usepackage[T1]{fontenc}

\usepackage[utf8]{inputenc}

\usepackage{microtype}

\usepackage{inconsolata}

\usepackage{graphicx}
\usepackage{booktabs}
\usepackage{multirow}
\usepackage{adjustbox}

\usepackage[linesnumbered,ruled,vlined]{algorithm2e}

\usepackage{amsmath,amsfonts,amssymb,amscd,xspace}
\SetKwInput{KwInput}{Input}                
\SetKwInput{KwOutput}{Output}

\title{Pre-Act: Multi-Step Planning and Reasoning Improves Acting in LLM Agents}

\author{
 \textbf{Mrinal Rawat\textsuperscript{1}},
 \textbf{Ambuje Gupta\textsuperscript{1}},
 \textbf{Rushil Goomer\textsuperscript{1}},
 \textbf{Alessandro Di Bari\textsuperscript{1}},
\\
 \textbf{Neha Gupta\textsuperscript{1}},
 \textbf{Roberto Pieraccini\textsuperscript{1}}
\\
 \textsuperscript{1}Uniphore
\\
 \small{
   \textbf{Correspondence:} \href{mailto:rawatmrinal06@gmail.com}{rawatmrinal06@gmail.com}
 }
}

\begin{document}
\maketitle
\begin{abstract}

The ReAct (Reasoning + Action) capability in large language models (LLMs) has become the foundation of modern agentic systems. Recent LLMs, such as DeepSeek-R1 and OpenAI o1/o3, exemplify this by emphasizing reasoning through the generation of ample intermediate tokens, which help build a strong premise before producing the final output tokens. In this paper, we introduce Pre-Act, a novel approach that enhances the agent's performance by creating a multi-step execution plan along with the detailed reasoning for the given user input. This plan incrementally incorporates previous steps and tool outputs, refining itself after each step execution until the final response is obtained. Our approach is applicable to both conversational and non-conversational agents. To measure the performance of task-oriented agents comprehensively, we propose a two-level evaluation framework: (1) turn level and (2) end-to-end. Our turn-level evaluation, averaged across five models, shows that our approach, Pre-Act, outperforms ReAct by 70\% in Action Recall on the Almita dataset. While this approach is effective for larger models, smaller models crucial for practical applications, where latency and cost are key constraints, often struggle with complex reasoning tasks required for agentic systems. To address this limitation, we fine-tune relatively small models such as Llama 3.1 (8B \& 70B) using the proposed Pre-Act approach. Our experiments show that the fine-tuned 70B model outperforms GPT-4, achieving a 69.5\% improvement in action accuracy (turn-level) and a 28\% improvement in goal completion rate (end-to-end) on the Almita (out-of-domain) dataset.

\end{abstract}

\section{Introduction}
Management of dialog systems has been a key focus of research in human-machine interaction. Various models have emerged, broadly categorized into distinct approaches \cite{pieraccini-huerta-2005-go}. Although inference and reinforcement-learning-based methods \cite{levin97_eurospeech} have been explored extensively, commercial applications have favored pragmatic solutions. Traditional rules-based or call-flow-driven systems \cite{10.1145/258549.258592} rely on predefined state transitions and procedural definitions. Although robust, they impose rigid dialog structures and require extensive manual design, leading to prolonged development cycles and high deployment costs, limiting scalability.

With the emergence of LLM-driven agents, dialog systems have evolved beyond static rule-based flows. LLMs enable open-ended interactions where business users are not constrained by predefined prompts, fostering more flexible, goal-oriented conversations. Agentic AI \cite{huang2024agent} acts autonomously, making decisions using advanced machine learning. However, effectiveness hinges on the underlying orchestrator, typically an LLM requiring strong reasoning and function-calling abilities. Recent models like DeepSeek-R1 \cite{deepseekai2025deepseekr1incentivizingreasoningcapability} and OpenAI o1/o3 generate additional reasoning-focused tokens and demonstrate these abilities, which is why approaches such as ReAct \cite{yao2023react} have become the primary foundation upon which today's agentic frameworks are built.

While ReAct provides a strong foundation for building agents, previous implementations had limitations. The reasoning component (generated as ``thought") typically focuses only on the reasoning required for the immediate action, making it inadequate for handling complex tasks that require executing a sequence of actions. Although the LLM is the system's cognitive backbone that drives reasoning, decision-making, and tool use, agent performance declines if the LLM orchestrator lacks intelligence or contextual awareness. Unfortunately, advanced reasoning remains largely restricted to proprietary models such as GPT-4 and Claude, limiting broader adoption.

To address these limitations, we make the following contributions in this paper:
\begin{itemize}
    \item Pre-Act, an enhanced version of ReAct that improves agents performance by generating a multi-step plan along with detailed reasoning for each action for the given task. Steps are executed sequentially, incorporating past actions and observations as context, refining the plan until the final output is obtained.
    
    \vspace{-0.2cm}
    \item A fine-tuning strategy leveraging datasets adapted for Pre-Act, enabling smaller models (e.g., Llama 8B, 70B) to match or surpass proprietary LLMs up to 20 times larger while reducing latency and cost, a critical aspect for real-world agentic applications.
    
    \vspace{-0.2cm}
    \item A two-level evaluation: (a) turn-level, assessing whether predicted actions align with ground truth and (b) end-to-end, measuring goal completion and progress rate.
    
\end{itemize}

\begin{figure*}[ht]
\centering
     \includegraphics[trim={2.1cm 5.5cm 2cm 4.9cm},clip,scale=0.68]{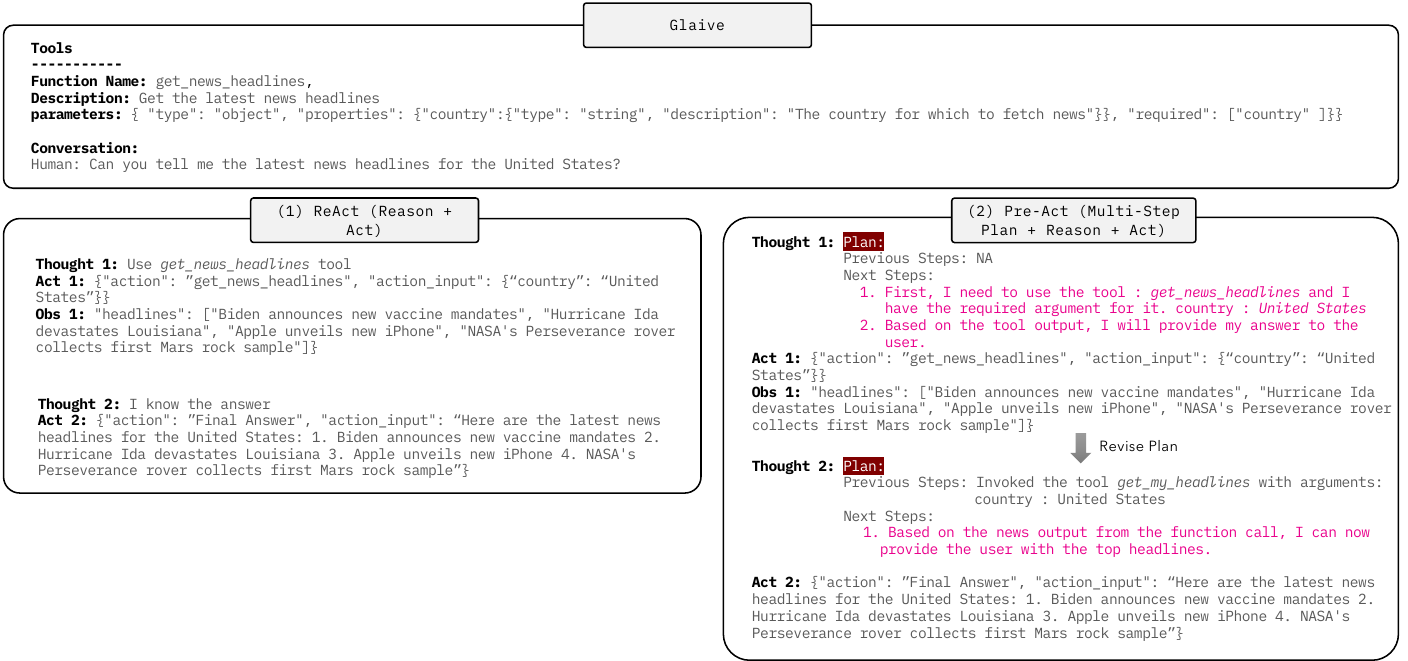}
      \caption{Side-by-side comparison of ReAct and Pre-Act (our approach) for an example use-case from the Glaive dataset. While this example demonstrates a single tool call, the approach can be extended to multiple tool calls.}
       \label{fig:refine_training}
\end{figure*}

\section{Related Work}
The concept of autonomous agents and agentic architectures is not new. The Open Agent Architecture \cite{article}, developed at SRI in the 1990s, enabled service creation through distributed autonomous agents coordinated by one or more facilitators. The Galaxy Architecture \cite{seneff98b_icslp}, initially developed at MIT for spoken dialogue systems, became the reference for the DARPA Communicator program, using a central HUB to exchange messages with multiple specialized servers.

Recent work has explored various strategies to enhance LLM reasoning and execution through advanced prompting techniques. Chain-of-Thought (CoT) prompting \cite{DBLP:journals/corr/abs-2201-11903} enables LLMs to break down complex problems into intermediate steps, while Self-Consistency \cite{wang2023selfconsistencyimproveschainthought} improves reliability by generating multiple reasoning paths. ReAct \cite{yao2023react} integrates reasoning with action execution, further extended by Tree-of-Thought (ToT) \cite{yao2023tree} and Graph-of-Thought \cite{besta2024got}, which introduce structured reasoning paths. Other approaches, such as Least-to-Most prompting \cite{zhou2023leasttomostpromptingenablescomplex} and Chain-of-Verification \cite{dhuliawala2023chainofverificationreduceshallucinationlarge}, enhance efficiency and reliability. However, these methods often generate redundant tokens and struggle with coherence across multiple steps, particularly in smaller models where computational efficiency is critical.

Fine-tuning is emerging as an effective method for achieving strong performance while using fewer tokens in the prompt \cite{ye2025limoreasoning}. Techniques such as instruction tuning \cite{wei2022finetuned} and RLHF \cite{10.5555/3600270.3602281}, when applied to smaller models with limited, but high-quality data, have shown promise in enhancing reasoning capabilities. However, these methods generally require datasets with explicit planning mechanisms to effectively learn structured reasoning and function-calling abilities, which are not readily available to the public and are difficult to obtain.

In this work, we address these limitations by introducing structured plans along with detailed reasoning generated by LLM. While prior work has explored planning in LLMs \cite{fu2023complexitybasedpromptingmultistepreasoning}, our approach extends it by generating a detailed multi-step plan with reasoning for each step, applicable to both conversational  and non-conversational AI agents. Additionally, existing evaluation frameworks for LLM agents often assess either individual action accuracy \cite{almita2024} or overall task performance \cite{gioacchini2024agentquestmodularbenchmarkframework}, but not both. Our two-level evaluation framework provides a more comprehensive assessment, considering both granular actions and overall task success.

\section{Approach}
\subsection{Incremental Multi-step planning}
\label{subsec:planning}
While the ReAct approach has emerged as a fundamental paradigm for building LLM-based agents, its current implementation typically focuses on single-step reasoning and (immediate) action generation \cite{yao2023react}. For each input request, the agent iteratively performs up to $n$ tool calls $[0,n]$ before arriving at a final answer. Despite its effectiveness, this approach often struggles with complex tasks that require long-term planning and sequential decision-making. To address this limitation, we propose Pre-Act, which generates a comprehensive multi-step plan. Instead of generating isolated reasoning steps, our approach formulates a structured execution plan with the reasoning that not only integrates previously executed steps, but also outlines the forthcoming steps necessary to achieve the final goal. 

For a given input request, our approach generates a plan consisting of $n+1$ sequential steps $S$ \{$s_1,s_2, \ldots, s_n, s_{fa}$\}, where each step $s_i$ explicitly specifies the intended action with a detailed thought, and the last step $s_{fa}$ represents the final answer.
When a step $s_i$ involves a tool call, the corresponding action $a_i$ is executed, and the resulting observation $o_i$ is incorporated into the reasoning process. For the subsequent step $s_{i+1}$, the system leverages both the current observation $o_i$ and the accumulated context $C_t=$ \{$(a_1,o_1), \ldots (a_i,o_i)$\}, to refine the plan and generate the next action, enhancing decision-making efficiency (see Figure \ref{fig:refine_training} for an example). This refinement process is particularly valuable when the outcome of a previous step deviates from expectations or results in failure, as it allows the agent to adapt its strategy dynamically. This iterative incorporation of past actions and observations augments the context $C_t$, ensuring that the agent maintains coherence and adaptability across multi-step interactions.  For the input prompt template, please refer to Appendix \ref{sec:react-plus-prompt}.

\subsection{Curriculum learning : Agents}
To train the models for agents, we adopt a curriculum learning \cite{10.1145/1553374.1553380} approach through incremental fine-tuning. Our process consists of two key stages:
\vspace{-0.09cm}

\begin{itemize}
    \item \textbf{Initial Fine-tuning:} We first fine-tune the pretrained Llama models on the Glaive dataset using the ReAct approach with minimal reasoning. This decision was pragmatic as the dataset is extensive and adding comprehensive reasoning annotations (as required for Pre-Act) would be very expensive. At this stage, our primary goal is to enable the model to perform agentic tasks specifically, learning to distinguish when to make tool calls with appropriate parameters versus when and what to generate as a response.
    \vspace{-0.15cm}
\item  \textbf{Progressive Refinement:} Using the trained checkpoint from the first stage, we further fine-tune the model on our proprietary dataset using the Pre-Act approach while preserving learning from the previous step through LoRA training, which modifies only a small fraction of parameters \cite{wistuba2023continuallearninglowrank}. For this smaller and high-quality dataset, we leverage our organization's annotation team to generate reasoning for each step. Importantly, the sequence of steps (multi-step plan) leading to the final answer was derived from the dataset itself; only the reasoning for each step was provided by annotators.
\end{itemize}

\subsubsection{Dataset for Fine-Tuning}
\label{sec:datasets}
Training models for conversational agentic capabilities requires datasets with two critical components: multi-turn dialogues and function (tool) calls with their corresponding responses. Such datasets are rare, and existing ones are not directly compatible with Pre-Act training requirements. The following section details our methodology for transforming the two datasets we used to train the models, ensuring alignment with our format requirements.

\textbf{Glaive Dataset}
We primarily leveraged the Glaive Function-Calling v2 dataset\footnote{\href{https://huggingface.co/datasets/glaiveai/glaive-function-calling-v2}{glaive-function-calling-v2}}, which covers conversational use-cases across multiple domains. Each data point comprises system instructions, tool definitions, and chat history. The chat history consists of user-assistant interactions, where assistant responses may incorporate function calls (FC) along with their corresponding function responses (FR). Formally, each conversation turn follows the structure: \{\texttt{[USER],[($FC_1, FR_1$)\dots ($FC_n, FR_n$)]\\,[ASSISTANT]}\}
where $FC_i$ represents a function call and $FR_i$ denotes its response, ($FC_n, FR_n$)  are optional depending on the specific interaction context. Algorithm \ref{algo} describes the transformation of each conversation turn into appropriate input-output pairs based on whether they contain tool calls or not. The sample output for both cases can be found in Appendix \ref{sec:appendix-glaive-io}.

\begin{algorithm}[th]
\DontPrintSemicolon
\KwInput{Conversation Turns $T = \{\text{USER}, [ (FC_1, FR_1), \dots, (FC_n, FR_n) ],$\\$\text{ASSISTANT} \}$, Instruction $I$, Tools Definition $TD$}

\SetKwFunction{SearchF}{Evaluation}
\SetKwProg{Fn}{Function}{:}{}
$D \gets \emptyset$; \tcp*{Dataset}
\ForEach{$t_i \in T$}{ 
    \If{$FC \notin t_i$}{
        $Ip_i \gets \{ I, TD, t_i(\text{USER}) \}$ \\ 
        $Op_i \gets$ Thought: I know the final answer. Action : $t_i(\text{ASSISTANT})$ \\ \tcp*{ASSISTANT response}
        $D \gets D \cup \{ (Ip_i, Op_i) \}$
    }
    \Else{
        $D_i \gets \emptyset;$\\
        \For{$j \in \{1, \dots, n\}$}{ \tcp*{N function calls}
            $Ip_j \gets \{ I, TD, t_i(\text{USER}), C((a_1,o_1),\dots,$\\$(a_{j-1}, o_{j-1})) \}$ \\ \tcp*{C is the accumulated context till $j-1$}
            $Op_j \gets$ Thought: Need to invoke tool : $FC_j$ \\
            Action : $FC_j$(arguments)\\ 
            $D_i \gets D_i \cup \{ (Ip_j, Op_j) \}$
        }
        $Ip_i \gets \{ I, TD, t_i(\text{USER}), C \}$ \\
        $Op_i \gets$ Thought: I know the final answer. Action : $t_i(\text{ASSISTANT})$ \\
        $D_i \gets D_i \cup \{ (Ip_i, Op_i) \}$\\
        $D \gets D \cup D_i$
    }
}

\caption{Pseudo-code for transforming the dataset to follow ReAct}
\label{algo}
\end{algorithm}

\textbf{Proprietary Dataset}
While the Glaive dataset provides a foundation for basic agentic capabilities, it has several limitations: minimal conversational deviations, lack of exception handling, and at most one consecutive tool call. To train more robust models, we curated a proprietary dataset spanning over 100 use-cases across domains including healthcare, manufacturing, telecommunications, banking, and finance. This dataset introduces complex scenarios that are absent in Glaive. Although the raw dataset also follows the same turn-based format, we specifically adapted it for Pre-Act using the methodology outlined in Section~\ref{subsec:planning}. Unlike Glaive, where we provide minimal reasoning (extractable from the dataset) for the tool calls and final answer, Pre-Act requires explicit reasoning at each step. Extending Algorithm \ref{algo}, we structured the dataset so that at each turn, the subsequent steps (i.e., tool calls) leading to the final answer could be derived. We incorporated placeholders for reasoning at each step, which were then completed by expert annotators. These annotators provided detailed explanations for every decision, ensuring a clear rationale behind tool calls and final responses (example in Fig.~\ref{fig:react_plus_sample}).


\section{Evaluation}
\label{sec:eval}
\subsection{Dataset}
We conducted our experiments on three datasets: Glaive, our proprietary dataset, and Almita. For Almita, we apply the same transformation as described in the previous section (annotated by our team for Pre-Act). The statistics of these datasets relevant for Level-1 evaluation are presented in Table \ref{tab:stats}. For Level 2 evaluation (end-to-end testing), we use the Almita dataset. Unlike Level 1, this stage requires running actual conversations with tool calling, needing an environment with tool implementations. Hence, we conduct our end-to-end evaluation on five complex Almita use-cases, illustrated in Table \ref{tab:e2e}. We selected these from 18 available cases by manually filtering out similar ones and those lacking tools or workflow information. Our contributions to this dataset, along with results and simulated conversations, are available in our Github repository.\footnote{\href{https://github.com/acl2025-submission/acl2025}{https://github.com/acl2025-submission/acl2025}}

\begin{table}[h]
\centering
\resizebox{1\columnwidth}{!}{
\begin{tabular}{llllll}
\hline
Dataset & \# Train & \# Val & \# Test & \#Use-cases & \#Tools\\ \hline
Glaive   & 341200        & 2672    & 4021          & 6702 & 1060    \\
Proprietary & 1852         & 128     & 378          & 119 & 360    \\
Almita & -         & -     & 1100           & 18 & 70    \\ \hline
\end{tabular}
}
\caption{Statistics of the three datasets used in Level-1 evaluation describing the number of instances used in training, validation, and test sets, respectively, along with total use-cases and tools.}
\label{tab:stats}
\end{table} 

\vspace{-0.1cm}
\subsection{Level 1: Turn Level Evaluation}
At this level, we evaluate each conversation turn for correctness against the ground truth (g.t.). For each given request, the action prediction from the LLM is either a Final Answer or a tool call. From a metrics perspective, we first calculate action recall. If the g.t. action is a tool call, we measure the F1 score and parameter match (full). If it is a Final Answer, we compute the F1 score and use a similarity model \cite{bge_embedding} to assess similarity to the g.t..

\begin{table*}[]
\resizebox{\textwidth}{!}{
\renewcommand{\arraystretch}{1.3} 

\begin{tabular}{|l|l|llllllllll|lllll|}
\hline
\multicolumn{1}{|c|}{\multirow{4}{*}{Models}}  & \multicolumn{1}{|c|}{\multirow{4}{*}{Approach}}                             & \multicolumn{10}{c|}{\textbf{In-Domain Test Dataset}}                                                   & \multicolumn{5}{c|}{\textbf{Out-of-Domain Dataset}}                                                        \\ \cline{3-17} 
\multicolumn{1}{|c|}{}  & \multicolumn{1}{|c|}{}                                                      & \multicolumn{5}{c|}{Glaive}                             & \multicolumn{5}{c|}{Proprietary}                          & \multicolumn{5}{c|}{Almita}                   \\ \cline{3-17} 
\multicolumn{1}{|c|}{} & \multicolumn{1}{|c|}{}                                                      & \multicolumn{1}{c|}{\multirow{2}{*}{\begin{tabular}[c]{@{}c@{}}\\Action\\ Recall\end{tabular}}} & \multicolumn{2}{c|}{Tool}                                                                                      & \multicolumn{2}{c|}{Final Answer}                                           & \multicolumn{1}{c|}{\multirow{2}{*}{\begin{tabular}[c]{@{}c@{}}\\Action\\ Recall\end{tabular}}} & \multicolumn{2}{c|}{Tool}                                                                                      & \multicolumn{2}{c|}{Final Answer}                                & \multicolumn{1}{c|}{\multirow{2}{*}{\begin{tabular}[c]{@{}c@{}}\\Action\\ Recall\end{tabular}}} & \multicolumn{2}{c|}{Tool}                                                                                      & \multicolumn{2}{c|}{Final Answer}                                \\ \cline{4-7} \cline{9-12} \cline{14-17} 
\multicolumn{1}{|c|}{}         & \multicolumn{1}{c|}{}                                              & \multicolumn{1}{c|}{}                                                                         & \multicolumn{1}{c|}{F1}              & \multicolumn{1}{c|}{\begin{tabular}[c]{@{}c@{}}Params\\ Match\\(Full)\end{tabular}} & \multicolumn{1}{c|}{F1}              & \multicolumn{1}{c|}{Sim.}            & \multicolumn{1}{c|}{}                                                                         & \multicolumn{1}{c|}{F1}              & \multicolumn{1}{c|}{\begin{tabular}[c]{@{}c@{}}Params\\ Match\\(Full)\end{tabular}} & \multicolumn{1}{c|}{F1}              & \multicolumn{1}{c|}{Sim.} & \multicolumn{1}{c|}{}                                                                         & \multicolumn{1}{c|}{F1}              & \multicolumn{1}{c|}{\begin{tabular}[c]{@{}c@{}}Params\\ Match\\(Full)\end{tabular}} & \multicolumn{1}{c|}{F1}              & \multicolumn{1}{c|}{Sim.} \\ \hline
\multirow{2}{*}{\begin{tabular}[c]{@{}l@{}}llama3.1 8B\\ (van.)\end{tabular}}      & \multicolumn{1}{l|}{ReAct}        & \multicolumn{1}{l|}{0.6064}                                                                   & \multicolumn{1}{l|}{0.5447}          & \multicolumn{1}{l|}{0.7868}                                                 & \multicolumn{1}{l|}{0.6532}          & \multicolumn{1}{l|}{0.8586}   & \multicolumn{1}{l|}{0.1528}                                                                   & \multicolumn{1}{l|}{0.1290}          & \multicolumn{1}{l|}{0.1556}                                                 & \multicolumn{1}{l|}{0.1903}          & \textbf{\underline{0.9413}}                    & \multicolumn{1}{l|}{0.1537}                                                                   & \multicolumn{1}{l|}{0.1906}          & \multicolumn{1}{l|}{0.2477}                                                 & \multicolumn{1}{l|}{0.1101}          &         0.6277            \\ \cline{2-17} 
  \multicolumn{1}{|l|}{}   & \multicolumn{1}{l|}{Pre-Act}          & \multicolumn{1}{l|}{-}                                                                   & \multicolumn{1}{l|}{-}          & \multicolumn{1}{l|}{-}                                                 & \multicolumn{1}{l|}{-}          & \multicolumn{1}{l|}{-}               & \multicolumn{1}{l|}{\textbf{0.4613}}                                                                   & \multicolumn{1}{l|}{\textbf{0.5087}}          & \multicolumn{1}{l|}{\textbf{0.4967}}                                                 & \multicolumn{1}{l|}{\textbf{0.3697}}          & 0.9324                    & \multicolumn{1}{l|}{\textbf{0.2779}}                                                                   & \multicolumn{1}{l|}{\textbf{0.3702}}          & \multicolumn{1}{l|}{\textbf{0.7216}}                                                 & \multicolumn{1}{l|}{\textbf{0.1293}}          & \textbf{0.7239}                   \\ \hline \hline
\multirow{2}{*}{\begin{tabular}[c]{@{}l@{}}llama3.1 70B\\(van.)\end{tabular}}    & \multicolumn{1}{l|}{ReAct}        & \multicolumn{1}{l|}{0.7771}                                                                   & \multicolumn{1}{l|}{0.767}           & \multicolumn{1}{l|}{0.7995}                                                 & \multicolumn{1}{l|}{0.7859}          & \multicolumn{1}{l|}{0.8971}            & \multicolumn{1}{l|}{0.3791}                                                                   & \multicolumn{1}{l|}{0.4214}          & \multicolumn{1}{l|}{0.5269}                                                 & \multicolumn{1}{l|}{0.3314}          & \textbf{0.9395}                    & \multicolumn{1}{l|}{0.3268}                                                                   & \multicolumn{1}{l|}{0.3643}          & \multicolumn{1}{l|}{0.5677}                                                 & \multicolumn{1}{l|}{0.2724}          &  0.7853         \\\cline{2-17}
\multicolumn{1}{|l|}{}   & \multicolumn{1}{l|}{Pre-Act}           &   \multicolumn{1}{l|}{-}                                                                   & \multicolumn{1}{l|}{-}           & \multicolumn{1}{l|}{-}                                                 & \multicolumn{1}{l|}{-}          & \multicolumn{1}{l|}{-}      & \multicolumn{1}{l|}{\textbf{0.5472}}                                                                   & \multicolumn{1}{l|}{\textbf{0.5609}}          & \multicolumn{1}{l|}{\textbf{0.5677}}                                                 & \multicolumn{1}{l|}{\textbf{0.5246}}          & 0.9137                    & \multicolumn{1}{l|}{\textbf{0.4045}}                                                                   & \multicolumn{1}{l|}{\textbf{0.4204}}          & \multicolumn{1}{l|}{\textbf{0.7508}}                                                 & \multicolumn{1}{l|}{\textbf{0.3846}}          & \textbf{0.8842}          \\ \hline \hline

\multirow{2}{*}{\begin{tabular}[c]{@{}l@{}}Nvidia Nemotron\\70B(van.)\end{tabular}}  & \multicolumn{1}{l|}{ReAct}              &  \multicolumn{1}{l|}{0.6771}                                                                   & \multicolumn{1}{l|}{0.6050}          & \multicolumn{1}{l|}{0.6578}                                                 & \multicolumn{1}{l|}{0.7478}          & \multicolumn{1}{l|}{0.8878}          & \multicolumn{1}{l|}{0.2377}                                                                   & \multicolumn{1}{l|}{0.1988}          & \multicolumn{1}{l|}{0.2258}                                                 & \multicolumn{1}{l|}{0.3002}          & 0.8013                    & \multicolumn{1}{l|}{0.1460}                                                                   & \multicolumn{1}{l|}{0.0647}          & \multicolumn{1}{l|}{0.1259}                                                 & \multicolumn{1}{l|}{0.2659}          &       0.7673             \\ \cline{2-17}
\multicolumn{1}{|l|}{}  & \multicolumn{1}{l|}{Pre-Act}              & \multicolumn{1}{l|}{-}                                                                   & \multicolumn{1}{l|}{-}          & \multicolumn{1}{l|}{-}                                                 & \multicolumn{1}{l|}{-}          & \multicolumn{1}{l|}{-}         & \multicolumn{1}{l|}{\textbf{0.4928}}                                                                   & \multicolumn{1}{l|}{\textbf{0.4966}}          & \multicolumn{1}{l|}{\textbf{0.4774}}                                                 & \multicolumn{1}{l|}{\textbf{0.4864}}          & \textbf{0.8955}                    & \multicolumn{1}{l|}{\textbf{0.3908}}                                                                   & \multicolumn{1}{l|}{\textbf{0.4049}}          & \multicolumn{1}{l|}{\textbf{0.7443}}                                                 & \multicolumn{1}{l|}{\textbf{0.3728}}          & \textbf{0.8370}                    \\ \hline \hline
\multirow{2}{*}{\begin{tabular}[c]{@{}l@{}}Deepseek-distil\\llama-3.1-70B(van.)\end{tabular}} & \multicolumn{1}{l|}{ReAct}  &   \multicolumn{1}{l|}{0.7823}                                                                   & \multicolumn{1}{l|}{0.7611}          & \multicolumn{1}{l|}{0.7314}                                                 & \multicolumn{1}{l|}{0.8004}          & \multicolumn{1}{l|}{0.9040}           & \multicolumn{1}{l|}{0.2756}                                                                   & \multicolumn{1}{l|}{0.2339}          & \multicolumn{1}{l|}{0.2168}                                                 & \multicolumn{1}{l|}{0.3555}          & 0.6965                    & \multicolumn{1}{l|}{0.3232}                                                                   & \multicolumn{1}{l|}{0.3725}          & \multicolumn{1}{l|}{0.6586}                                                 & \multicolumn{1}{l|}{0.2511}          &        \textbf{0.8438}             \\ \cline{2-17}
\multicolumn{1}{|l|}{} & \multicolumn{1}{l|}{Pre-Act}  &\multicolumn{1}{l|}{-}                                                                   & \multicolumn{1}{l|}{-}          & \multicolumn{1}{l|}{-}                                                 & \multicolumn{1}{l|}{-}          & \multicolumn{1}{l|}{-}        & \multicolumn{1}{l|}{\textbf{0.6531}}                                                                   & \multicolumn{1}{l|}{\textbf{0.6712}}          & \multicolumn{1}{l|}{\textbf{0.4421}}                                                 & \multicolumn{1}{l|}{\textbf{0.6435}}          & \textbf{0.7157}                    & \multicolumn{1}{l|}{\textbf{0.5036}}                                                                   & \multicolumn{1}{l|}{\textbf{0.4697}}          & \multicolumn{1}{l|}{\textbf{0.7518}}                                                 & \multicolumn{1}{l|}{\textbf{0.5488}}          & 0.8418                    \\ \hline \hline
\multirow{2}{*}{\begin{tabular}[c]{@{}l@{}}gpt-4-turbo\end{tabular}}   & \multicolumn{1}{l|}{ReAct}                                                                    & \multicolumn{1}{l|}{0.9265}                                                                   & \multicolumn{1}{l|}{0.8649}          & \multicolumn{1}{l|}{0.9145}                                                 & \multicolumn{1}{l|}{0.9496}          & \multicolumn{1}{l|}{0.9449}            & \multicolumn{1}{l|}{0.4933}                                                                   & \multicolumn{1}{l|}{0.5150}          & \multicolumn{1}{l|}{0.5612}                                                 & \multicolumn{1}{l|}{0.5057}          & 0.8870                   & \multicolumn{1}{l|}{0.4430}                                                                   & \multicolumn{1}{l|}{0.3995}          & \multicolumn{1}{l|}{0.6634}                                                 & \multicolumn{1}{l|}{0.4930}          &       0.7452              \\ \cline{2-17}
\multicolumn{1}{|l|}{}    & \multicolumn{1}{l|}{Pre-Act}                                                                   & \multicolumn{1}{l|}{-}                                                                   & \multicolumn{1}{l|}{-}          & \multicolumn{1}{l|}{-}                                                 & \multicolumn{1}{l|}{-}          & \multicolumn{1}{l|}{-}        & \multicolumn{1}{l|}{\textbf{0.6131}}                                                                   & \multicolumn{1}{l|}{\textbf{0.6019}}          & \multicolumn{1}{l|}{\textbf{0.5870}}                                                 & \multicolumn{1}{l|}{\textbf{0.6293}}          & \textbf{0.9116}                    & \multicolumn{1}{l|}{\textbf{0.5449}}                                                                   & \multicolumn{1}{l|}{\textbf{0.4616}}          & \multicolumn{1}{l|}{\textbf{0.7532}}                                                 & \multicolumn{1}{l|}{\textbf{0.6214}}          & \textbf{0.8201}                    \\ \hline \hline
\begin{tabular}[c]{@{}l@{}}llama3.1-8B\\(f.t.)\end{tabular}  & \multicolumn{1}{l|}{Pre-Act}              & \multicolumn{1}{l|}{0.9881}                                                                   & \multicolumn{1}{l|}{0.9744}          & \multicolumn{1}{l|}{0.9289}                                                 & \multicolumn{1}{l|}{0.9922}          & \multicolumn{1}{l|}{0.9623}          & \multicolumn{1}{l|}{0.8911}                                                                   & \multicolumn{1}{l|}{0.8327}          & \multicolumn{1}{l|}{0.5806}                                                 & \multicolumn{1}{l|}{0.9363}          & 0.8894           & \multicolumn{1}{l|}{0.8706}                                                                   & \multicolumn{1}{l|}{0.7464}          & \multicolumn{1}{l|}{0.7475}                                                 & \multicolumn{1}{l|}{0.9306}          & 0.8256                    \\ \hline
\begin{tabular}[c]{@{}l@{}}llama3.1-70B\\(f.t.) \end{tabular}   & \multicolumn{1}{l|}{Pre-Act}            & \multicolumn{1}{l|}{\textbf{\underline{0.9929}}}                                                          & \multicolumn{1}{l|}{\textbf{\underline{0.9848}}} & \multicolumn{1}{l|}{\textbf{\underline{0.9586}}}                                        & \multicolumn{1}{l|}{\textbf{\underline{0.9954}}} & \multicolumn{1}{l|}{\textbf{\underline{0.9826}}} & \multicolumn{1}{l|}{\textbf{\underline{0.9111}}}                                                          & \multicolumn{1}{l|}{\textbf{\underline{0.8625}}} & \multicolumn{1}{l|}{\textbf{\underline{0.6258}}}                                        & \multicolumn{1}{l|}{\textbf{\underline{0.9523}}} & 0.8671                    & \multicolumn{1}{l|}{\textbf{\underline{0.9238}}}                                                          & \multicolumn{1}{l|}{\textbf{\underline{0.8636}}} & \multicolumn{1}{l|}{\textbf{\underline{0.7961}}}                                        & \multicolumn{1}{l|}{\textbf{\underline{0.9496}}} & \textbf{\underline{0.8861}}                    \\ \hline
\end{tabular}
}
\caption{Results on two in-domain test sets (Glaive and a proprietary dataset) and one out-of-domain set (Almita). "Van." refers to the vanilla (pretrained) model, while "f.t." denotes fine-tuned. Higher values indicate better performance. Bold numbers highlight the better approach between ReAct and Pre-Act, while bold and underlined numbers indicate the best overall model in each column.}
\label{tab:level1}
\end{table*}

\vspace{-0.1cm}
\subsection{Level 2: End-to-End (E2E) Evaluation}
A key limitation of the Level-1 evaluation framework is that it focuses solely on individual turns rather than evaluating the conversation or task as a whole. Most publicly available datasets  consist primarily of "happy path" scenarios, where users provide correct and straightforward information. However, real-world interactions are often more unpredictable, where users may deviate from topics, providing partial or incorrect information or errors introduced through automatic speech recognition (ASR) in voice interfaces. To address such scenarios and gain a comprehensive, end-to-end view of AI agent performance, we introduce the Level-2 evaluation that assesses an AI agent's ability to manage complex conversations within a business context, where the primary goal is to accomplish a predefined task following a particular workflow.

\subsubsection{Aspects of Level 2 Evaluation}
\textbf{Milestone Creation: }
We draw inspiration from AgentQuest \cite{gioacchini2024agentquestmodularbenchmarkframework}, which highlights the importance of breaking down tasks into mission-critical milestones and introduces metrics like progress rate and goal completion to measure an agent's task completion performance. However, in our setting, defining milestones and evaluating progress is not straightforward, as tasks can vary significantly. We utilize GPT-4 to create a structured set of incremental and/or conditional milestones that align with the user-defined workflow and tools (refer to Appendix \ref{sec:prompt-milestones} for the prompt). Additionally, we prompt it to establish dependencies between these milestones. The resulting output is then used to construct a milestone dependency graph. Since this automatically generated graph may not always be entirely accurate, we incorporate human verification and refinement to ensure its correctness. Once finalized, this graph is stored for that use-case and serves as an input for agent evaluation (sample provided in Appendix \ref{sec:sample_milestones}).

\textbf{Simulating Environments: }
    An effective approach to end-to-end evaluation of conversational systems is using a synthetic user to simulate interactions based on predefined scenarios, a concept introduced in \citet{levin2000stochastic} and later expanded in \citet{pietquin2006probabilistic}. To test our AI agents, we built a simulation environment leveraging GPT-4 as a synthetic user and let it interact with the AI agent. By dynamically assigning personas, we model user behaviors ranging from simple queries to complex ethical dilemmas, security challenges, and critical decision-making. This ensures rigorous assessment of the AI’s adherence to business workflows and handling of sensitive situations, enhancing its trustworthiness.


\textbf{Evaluating Goal Completion Metrics: }
We use LLM-as-a-judge \cite{10.5555/3666122.3668142} (GPT4) to evaluate simulated conversations. For goal completion, the LLM considers the entire conversation, the AI agent’s task (instructions), available tools, and then reviews stored milestones to verify whether they were achieved successfully in the correct sequence. To calculate the progress rate, we map the achieved milestones (predicted by the LLM) against the milestone dependency graph and measure progress by traversing from the start node through the completed milestones, calculating the ratio of the distance covered to the total distance from start to end. This provides a measure of how far the conversation has progressed toward the final goal. If 100\% of the milestones are completed to reach the end, the goal is considered fully achieved (prompt given in Appendix \ref{sec:prompt-e2e}).

\begin{table}[th!]
\centering
\resizebox{1\columnwidth}{!}{
\begin{tabular}{|l|l|llllllllll|}
\hline
\multirow{3}{*}{\textbf{Models}}                              & \multirow{3}{*}{\textbf{Approach}} & \multicolumn{10}{c|}{\textbf{Use-cases}}                                                                                                                                                                                                                                                                                                                                                                              \\ \cline{3-12} 
                                                              &                                    & \multicolumn{2}{l|}{\textbf{\begin{tabular}[c]{@{}l@{}}Order\\ Discrepancy\end{tabular}}} & \multicolumn{2}{l|}{\textbf{\begin{tabular}[c]{@{}l@{}}Internet\\ Ping\end{tabular}}} & \multicolumn{2}{l|}{\textbf{\begin{tabular}[c]{@{}l@{}}Gift\\ Card\end{tabular}}} & \multicolumn{2}{l|}{\textbf{\begin{tabular}[c]{@{}l@{}}Digital\\ Download\end{tabular}}} & \multicolumn{2}{l|}{\textbf{Delivery}}             \\ \cline{3-12} 
                                                              &                                    & \multicolumn{1}{l|}{\textbf{GC}}            & \multicolumn{1}{l|}{\textbf{PR}}            & \multicolumn{1}{l|}{\textbf{GC}}          & \multicolumn{1}{l|}{\textbf{PR}}          & \multicolumn{1}{l|}{\textbf{GC}}        & \multicolumn{1}{l|}{\textbf{PR}}        & \multicolumn{1}{l|}{\textbf{GC}}            & \multicolumn{1}{l|}{\textbf{PR}}           & \multicolumn{1}{l|}{\textbf{GC}}   & \textbf{PR}   \\ \hline
\multirow{2}{*}{gpt-4-turbo}                                   & ReAct                              & \multicolumn{1}{l|}{0.28}                   & \multicolumn{1}{l|}{0.39}                   & \multicolumn{1}{l|}{0.00}                 & \multicolumn{1}{l|}{0.11}                 & \multicolumn{1}{l|}{0.18}               & \multicolumn{1}{l|}{0.33}               & \multicolumn{1}{l|}{0.55}                   & \multicolumn{1}{l|}{0.79}                  & \multicolumn{1}{l|}{0.60} & 0.74 \\ \cline{2-12} 
                                                              & Pre-Act                             & \multicolumn{1}{l|}{0.63}                   & \multicolumn{1}{l|}{0.72}                   & \multicolumn{1}{l|}{0.52}                 & \multicolumn{1}{l|}{0.73}                 & \multicolumn{1}{l|}{0.75}               & \multicolumn{1}{l|}{0.74}               & \multicolumn{1}{l|}{0.68}                   & \multicolumn{1}{l|}{0.90}                  & \multicolumn{1}{l|}{\textbf{0.66}}          & \textbf{0.89}          \\ \hline
\begin{tabular}[c]{@{}l@{}}llama-3.1\\ 70B(f.t.)\end{tabular} & Pre-Act                             & \multicolumn{1}{l|}{\textbf{0.75}}          & \multicolumn{1}{l|}{\textbf{0.89}}          & \multicolumn{1}{l|}{\textbf{1.00}}        & \multicolumn{1}{l|}{\textbf{1.00}}        & \multicolumn{1}{l|}{\textbf{0.90}}      & \multicolumn{1}{l|}{\textbf{0.86}}      & \multicolumn{1}{l|}{\textbf{0.89}}          & \multicolumn{1}{l|}{\textbf{0.97}}         & \multicolumn{1}{l|}{0.60}          & 0.79          \\ \hline
\end{tabular}
}
\caption{End-to-End Evaluation results on Almita Dataset use-cases. GC represents Goal Completion and PR represents progress rate.}
\label{tab:e2e}
\end{table}

\section{Results and Discussion}
Table \ref{tab:level1} highlights two key findings: (a) the impact of Pre-Act on pretrained models (first five in the table) compared to ReAct, and (b) a comparison of fine-tuned models using Pre-Act against others. Pre-Act consistently outperforms ReAct across action recall, tool calls, and final answer similarity on both the proprietary and Almita datasets, with a minor drop in final answer similarity for Llama 3.1 8B \& 70B on the proprietary dataset. On average (not shown in the table), Pre-Act improves action recall over ReAct by \textbf{102\%} on the proprietary dataset and \textbf{70\%} on Almita. A comparison with Glaive was not possible due to missing Pre-Act annotations. 

The second part of our results highlights the impact of fine-tuning. On the in-domain test sets, our fine-tuned 70B model outperforms GPT-4 and its vanilla counterpart by 7.1\% and 27.7\% on the Glaive dataset. To assess generalization, we also evaluated on the Almita dataset (out-of-domain---applicable in case of f.t.), where our fine-tuned 70B model outperformed GPT-4 (Pre-Act) by 69.5\% and its vanilla counterpart by 128\% in action recall. The consistent performance gains both in-domain and out-of-domain underscore the robustness of our fine-tuning approach. Our findings highlight that smaller fine-tuned models can match or even surpass the performance of significantly larger proprietary systems, making them an efficient and scalable alternative.


Table \ref{tab:e2e} presents the end-to-end results for conversations generated synthetically. We ran the simulation 50 times across all five use-cases and reported the average goal completion (GC) rate and progress rate. As shown in Table \ref{tab:e2e}, our fine-tuned model, despite being significantly smaller in size, outperforms both the GPT with ReAct and Pre-Act approaches in most use-cases. On average (not shown in the table), our model achieves a GC rate of \textbf{0.82}, compared to \textbf{0.32} for GPT with ReAct and \textbf{0.64} for GPT with Pre-Act. It is noteworthy that the Pre-Act approach with GPT demonstrates significant improvements compared to GPT with ReAct, while our fine-tuned model achieves even higher GC and progress rates in challenging scenarios.

\section{Conclusion and Future Work}
\vspace{-0.12cm}
We have introduced Pre-Act, a method that improves LLM agent performance through multi-step planning and reasoning. We also proposed a fine-tuning strategy that allows smaller models to achieve performance comparable to larger LLMs with lower latency and cost, along with a two-level evaluation framework for rigorous assessment. In the future, we aim to enhance model robustness by incorporating complex scenarios and recovery paths into training data and adopting more deterministic evaluation methods to mitigate the volatility in LLM-as-a-judge assessments.

\bibliography{acl_latex}

\appendix

\begin{table*}[!th]
\centering
\begin{tabular}{|l|c|c|c|c|c|c|c|c|c|}
\hline
\multirow{3}{*}{\textbf{Models}} & \multicolumn{9}{c|}{\textbf{Glaive Dataset}}  \\ \cline{2-10} 
 & \multirow{2}{*}{\textbf{\begin{tabular}[c]{@{}c@{}}Action\\ Recall\end{tabular}}} 
 & \multicolumn{4}{c|}{\textbf{Function}} 
 & \multicolumn{4}{c|}{\textbf{Final Answer}} \\ \cline{3-10} 
 &  & \textbf{Recall} & \textbf{Precision} & \textbf{F1} & \textbf{\begin{tabular}[c]{@{}c@{}}Params\\ Match\\ (Full)\end{tabular}} 
 & \textbf{Recall} & \textbf{Precision} & \textbf{F1} & \textbf{Sim.} \\ \hline
\begin{tabular}[c]{@{}l@{}}Llama3.1\\ 8B- Step1\end{tabular}  
 & 0.9960 & 0.9946 & 0.9946 & 0.9946 & 0.9601 & 0.9968 & 0.9968 & 0.9968 & 0.9983 \\ \hline
\begin{tabular}[c]{@{}l@{}}Llama3.1\\ 8B- Step2\end{tabular}  
 & 0.9881 & 0.9793 & 0.9697 & 0.9744 & 0.9289 & 0.9908 & 0.9938 & 0.9922 & 0.9623 \\ \hline
\begin{tabular}[c]{@{}l@{}}Llama3.1\\ 70B- Step1\end{tabular}  
 & 0.9965 & 0.9932 & 0.9972 & 0.9951 & 0.9601 & 0.9984 & 0.9960 & 0.9971 & 0.9979 \\ \hline
\begin{tabular}[c]{@{}l@{}}Llama3.1\\ 70B- Step2\end{tabular}  
 & 0.9929 & 0.9964 & 0.9736 & 0.9848 & 0.9586 & 0.9919 & 0.9991 & 0.9954 & 0.9826 \\ \hline
\end{tabular}
\caption{Results on Glaive Dataset at Step-1 (Initial Fine-tuning) and Step-2 (Progressive refinement) to assess the impact of curriculum learning}
\label{tab:ct}
\end{table*}

\section{Training Details}
We used the TRL library \cite{vonwerra2022trl} from Hugging Face to train the Llama 3.1 8B and 70B models. Training was performed with the QLoRA \cite{dettmers2023qlora} technique for one epoch, using a per-device batch size of 1, gradient accumulation steps of 1, a learning rate of 5e-5, and a maximum sequence length of 4096.

\section{Example Output for Glaive Dataset}
\begin{figure}[ht!]
\centering

     \includegraphics[trim={2cm 8cm 16cm 4cm},clip,scale=0.8]{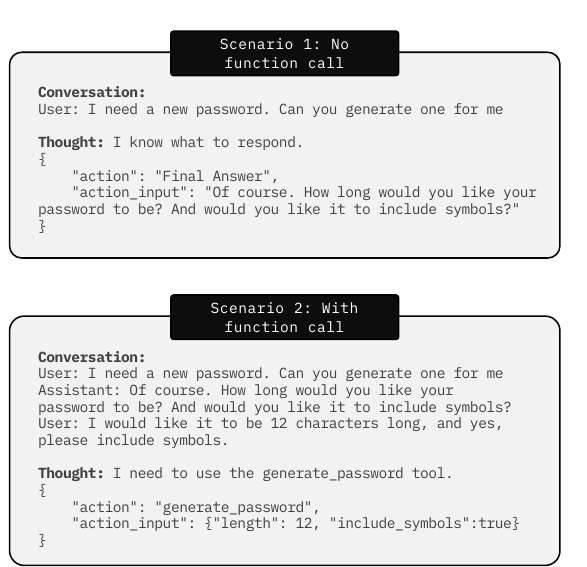}
      \caption{Sample Output for Glaive Dataset - ReAct}
       \label{fig:glaive-io}
\end{figure}

Please refer to Figure \ref{fig:glaive-io}.

\label{sec:appendix-glaive-io}

\section{Impact of Curriculum Learning}
We aim to study the impact of curriculum learning, as one potential issue with this approach is catastrophic forgetting. This occurs because learning happens in independent stages, meaning the model may forget earlier stages while training on later ones. In this work, however, we use instruction fine-tuning with LoRA, which modifies only a limited number of parameters in the network, thereby mitigating the risk of forgetting \cite{ren2024analyzingreducingcatastrophicforgetting}. To assess this impact, we evaluated metrics on the Glaive dataset at step 1 (where this dataset was used) and compared them to step 2 (where the model was trained on a different dataset) to determine whether it retained knowledge from step 1. Table \ref{tab:ct} shows that for both the fine-tuned models 8B and 70B, the degradation was minimal. Specifically, the 8B model experienced only a 0.80\% drop in action recall, while the 70B model had a mere 0.36\% decrease. A similar trend is observed across other metrics.

\section{Pre-Act Prompt Template}
\label{sec:react-plus-prompt}
\begin{figure*}[ht!]
\centering
     \includegraphics[trim={0.5cm 0.3cm 0.5cm 0.5cm},clip,scale=0.77]{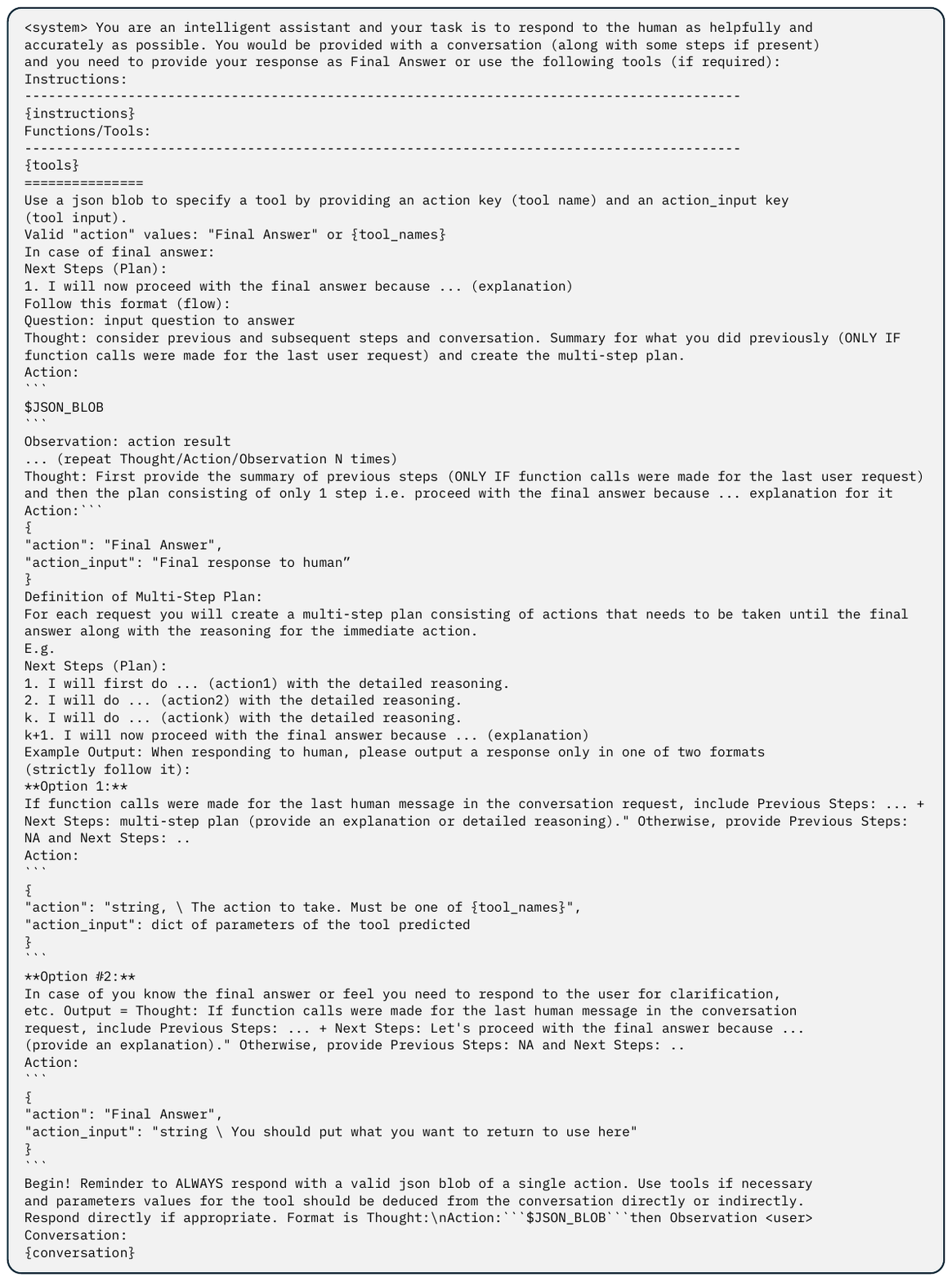}
      \caption{Prompt Template used for Pre-Act : Multi-Step Planning}
       \label{react_prompt}

\end{figure*}

The Pre-Act prompt template is designed to guide the AI agent through structured planning and action execution within a conversation. As shown in Figure \ref{react_prompt}, the template provides clear instructions on how to generate responses, make decisions, and utilize tools effectively. It ensures that the agent follows a systematic approach by maintaining context awareness and planning subsequent steps, thereby enhancing the overall interaction quality and task completion accuracy.

\section{Sample cases for Pre-Act}
\begin{figure}[!ht]
\centering
     \includegraphics[trim={2cm 12.5cm 16cm 4cm},clip,scale=0.8]{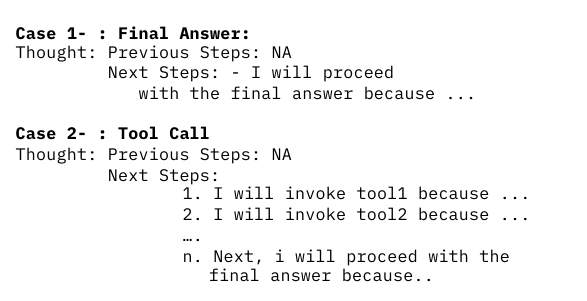}

\caption{Sample}
\label{fig:react_plus_sample}
\end{figure}

\section{Prompt Template for Milestones Creation}
\label{sec:prompt-milestones}

\begin{figure*}[ht!]
\centering

     \includegraphics[trim={0.5cm 0.5cm 0.5cm 0.4cm},clip,scale=0.75]{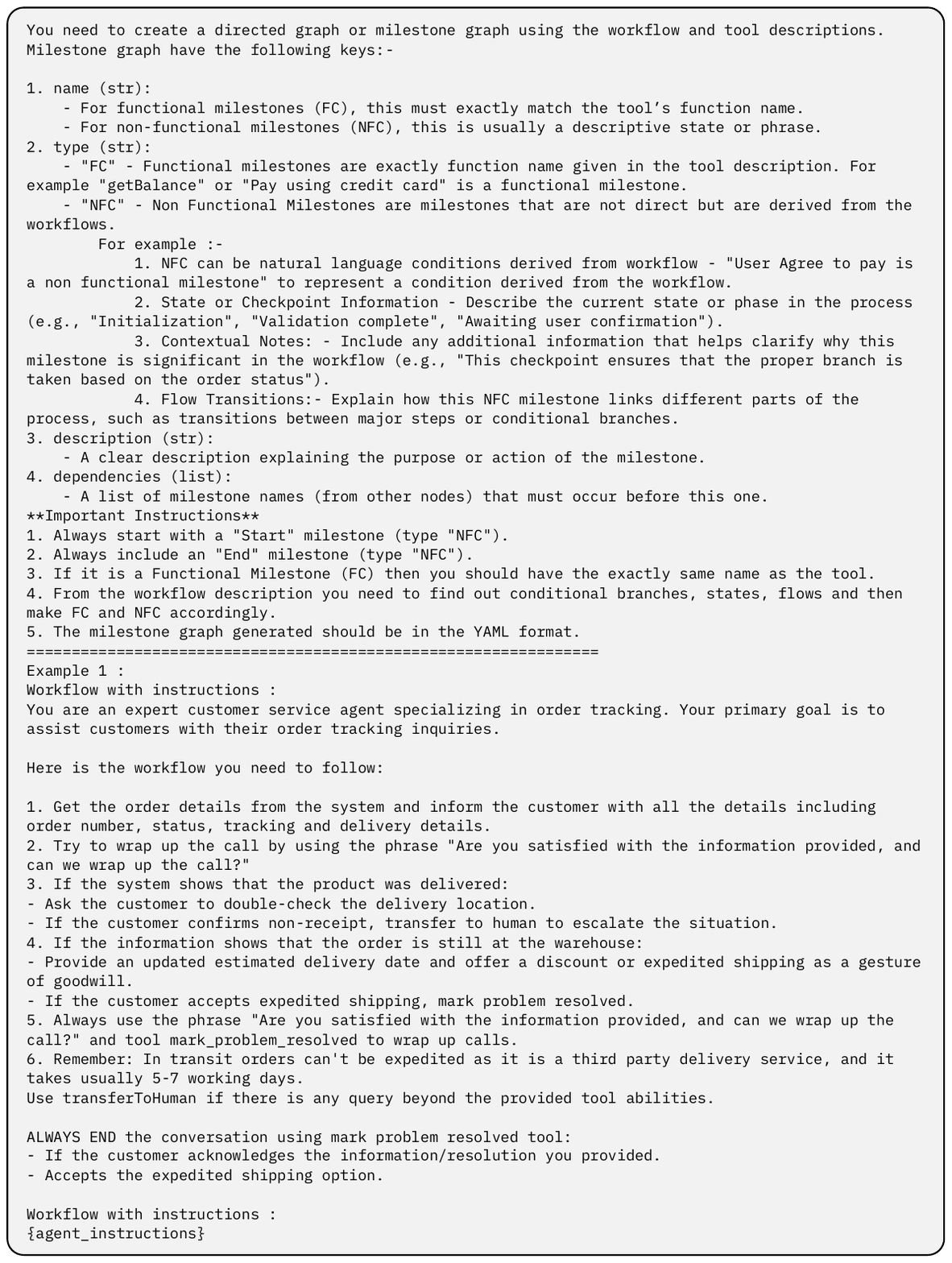}
    
      \caption{Prompt Template for Milestones Creation}
       \label{fig:prompt-milestones}

\end{figure*}

As shown in Figure \ref{fig:prompt-milestones}, the milestone graph is constructed by analyzing the workflow and identifying functional and non-functional milestones. The prompt template provides a structured approach to generating directed milestone graphs from workflow and tool descriptions. It outlines the essential components of a milestone, including the name, type (either functional or non-functional), description, and dependencies. The template emphasizes the importance of distinguishing between Functional Milestones (FC), which directly correspond to tool functions, and Non-Functional Milestones (NFC), which represent states, conditions, or contextual transitions within the workflow. It also provides clear guidelines for structuring the graph in YAML format, ensuring consistency and completeness.

\section{Sample Milestone Dependency Graph}
\label{sec:sample_milestones}

\begin{figure*}[ht!]
\centering
     \includegraphics[trim={0.2cm 0.2cm 0.1cm 0.2cm},clip,scale=0.35]{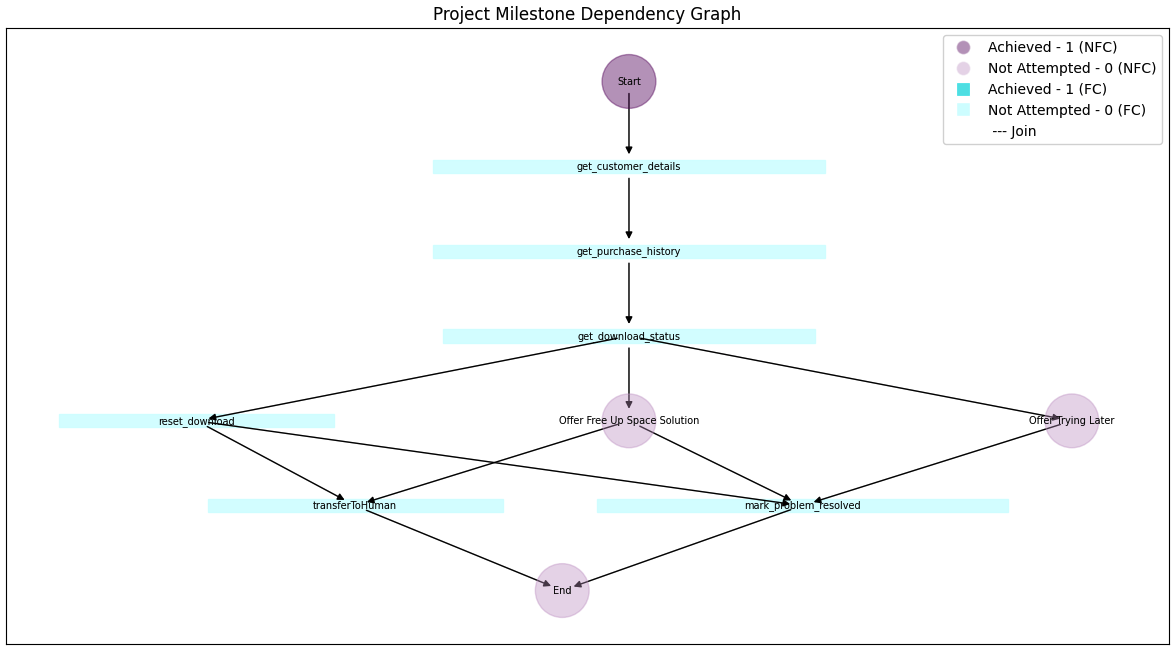}
      \caption{Sample Milestone Dependency Graph}
       \label{fig:milestone_png}

\end{figure*}

The Sample Milestone Dependency Graph shown in Figure \ref{fig:milestone_png}, visually represents the sequence and interconnections between various milestones within a workflow. It highlights both Functional Milestones (FC), which correspond to specific tool functions, and Non-Functional Milestones (NFC), representing conditions, states, or transitional checkpoints. The graph clearly delineates the dependencies between milestones, illustrating how the completion of one milestone enables the progression to the next. This structured representation aids in understanding the logical flow and ensures that all prerequisites are met before advancing to subsequent steps.

\section{Prompt Template for GPT-4 as Judge (E2E evaluation)}
\label{sec:prompt-e2e}
\begin{figure*}[ht!]
\centering
     \includegraphics[trim={0.5cm 0.5cm 0.5cm 0.4cm},clip,scale=0.75]{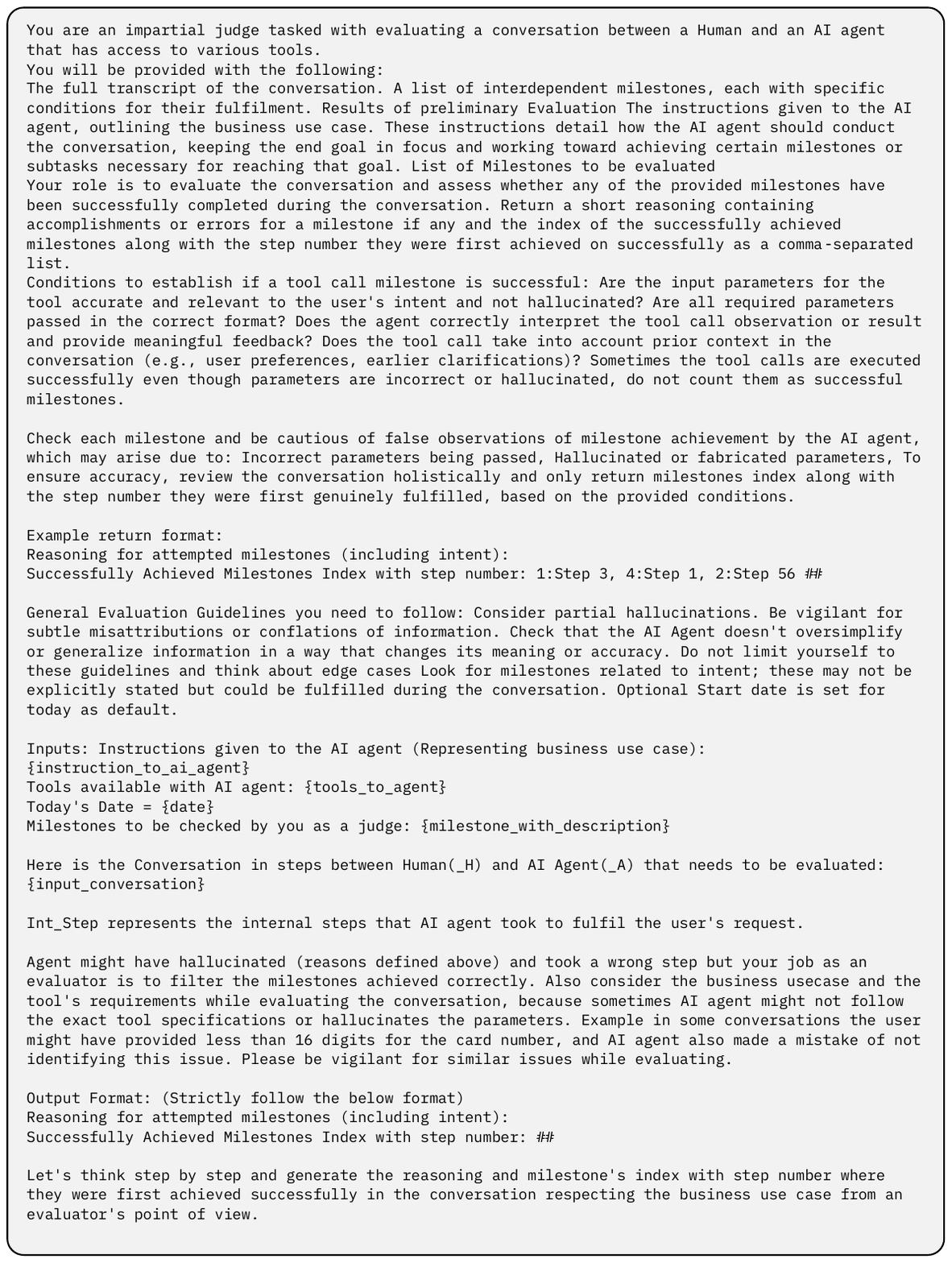}
      \caption{Prompt Template for GPT-4 as Judge (E2E evaluation)}
       \label{fig:prompt-e2e}

\end{figure*}

As shown in Figure \ref{fig:prompt-e2e} the prompt template contains a structured guideline for evaluating conversations between a human user and an AI agent with tool access. It outlines the evaluator's role, which is to assess whether the AI agent has successfully achieved specified milestones during the conversation. The template emphasizes accuracy and vigilance, cautioning against false positives, hallucinated data, or incomplete parameter validation. It also provides criteria for verifying successful tool calls and reasoning for attempted milestones. Additionally, it specifies a strict output format for presenting successfully achieved milestones along with the step numbers where they were first genuinely fulfilled

\end{document}